\documentclass{article}

\usepackage{arxiv}

\usepackage[utf8]{inputenc} 
\usepackage[T1]{fontenc}    
\usepackage{hyperref}       
\usepackage{url}            
\usepackage{booktabs}       
\usepackage{amsfonts}       
\usepackage{nicefrac}       
\usepackage{microtype}      
\usepackage{lipsum}
\usepackage{graphicx}

\title{PI-GAN: Learning Pose Independent representations for multiple profile face synthesis}

\author{
Hamed Alqahtani \\
  Macquarie University\\
  Sydney, Australia\\
  \texttt{hsqahtani@kku.edu.sa} \\
}

\begin{document}
\maketitle

\begin{abstract}
Generating a pose-invariant representation capable of synthesizing multiple face pose views from a single pose is still a difficult problem. The solution is demanded in various areas like multimedia security, computer vision, robotics, etc. Generative adversarial networks (GANs) have encoder-decoder structures possessing the capability to learn pose-independent representation incorporated with discriminator network for realistic face synthesis. We present PIGAN, a cyclic shared encoder-decoder framework, in an attempt to solve the problem. As compared to traditional GAN, it consists of secondary encoder-decoder framework sharing weights from the primary structure and reconstructs the face with the original pose. The primary framework focuses on creating disentangle representation, and secondary framework aims to restore the original face. We use CFP high-resolution, realistic dataset to check the performance.
\end{abstract}


\section{Introduction}
Generating photorealistic \cite{ledig2017photo} faces has several applications in the area of computer vision, security, person identification, face recognition, face synthesis, etc. Wide range of methods has been proposed in this area addressing the problem and challenges. One of the main challenges comes due to the pose-variation among the faces of the same identity \cite{chai2007locally}. This creates a need to devise a method robust to pose-variation. Deep learning has shown rapid progress working with images, but these data-driven approaches need ample amount of data to be trained on. Face recognition has made noticeable progress in an unconstrained environment because of available sufficient annotated sample data and data-driven deep learning approaches. Deep learning models already performs better than Human recognition, but it is still challenging for real-world application. Existing solutions approaching pose variation either use trained features\cite{Chen_2013,Schroff_2015} or synthesize new face from identity representation \cite{Zhu_2013,Zhu2014MultiViewPA,Tran2017DisentangledRL}.

Trained features often employ feature descriptors such as HAAR \cite{Viola} and LBP \cite{Ahonen_2006} to recognize pose variation. Synthesis methods use deep learning generative networks which try to capture identity information of each individual and reconstruct faces from the learned representation. Existing deep learning techniques suffers from large pose cases due to the trade-off between improving pose-discrimination and identity-discrimination\cite{ding2016comprehensive,alqahtani2019analysis,Alqahtani2019}.

Some 3D-model based methods create the general\cite{Hassner_2015,ding2015robust,li2012morphable} or idiosyncratic \cite{Taigman_2014,Xiangyu_Zhu_2015} 3D model and try to observe the correlation between the 2D image and the 3D model to generate the frontal face. Recently developed deep learning methods learn the identity and pose representation separately in latent space \cite{Zhu2014MultiViewPA,Tran2017DisentangledRL}. Although their results have already crossed benchmark still granular details are absent in the synthesized image. Lacking minor details can change the identity of how a person looks concerning age, skin texture, the position of mole, eyebrows, etc. This leads to incorrect face recognition. Also, these learned representations are heavily data-dependent, meaning their performance decrease while performing on out of sample data.
In PIGAN, secondary encoder-decoder is incorporated to improve minor details while reconstructing back the face with original pose and identity. The secondary framework provides a pixel-wise loss between the input image and synthesized image, which ideally should be exactly same\cite{kulkarni2015deep}. The loss adjusts the representation learned by the primary framework in the direction where minor details are enhanced. The discriminator is modified to perform pose and identity classification along with discriminating between real and fake faces.

\section{Related Work}

\subsubsection{Face synthesis}
Generating frontal face from a given non-frontal face (also known as face normalization) is a strenuous problem as image properties are primarily dependent on face profile \cite{hassner2015effective,kan2014stacked,sagonas2015robust,yim2015rotating,zhu2015high,zhu2014multi}. Traditional methods model the 3D mean for frontal view synthesis \cite{Hassner_2015,parkhi2015deep,schroff2015facenet}. Deep learning methods focus on preserving face information \cite{Zhu_2013}. Yang et al. \cite{Yang2015WeaklysupervisedDW} proposes a recurrent network to rotate faces via hidden units. Recently developed proposition generates face via manifold traversal \cite{Schroff_2015}. Following these research work, our framework assimilates identity preserving manifolds independent of viewpoint information.

\subsubsection{Generative adversarial networks}
Goodfellow et al. \cite{Goodfellow2014GenerativeAN} first introduced GANs to learn the target distribution of the given data and can generate a sample from the distribution which closes follows the properties of the distribution. Later, Conditional GAN by Mehdi Mirza and Simon Osindero \cite{Mirza2014ConditionalGA} presented a stable approach for target distribution learning. These are single pathway networks \cite{springenberg2015unsupervised,radford2015unsupervised} which can project the input data on a latent space and adjust the space to fit the distribution. In the case of face synthesis, the large manifold may lack finer details, which are necessary for best face recognition performance. We use two single pathway architecture back to back to generate an original image (ideally exact same), thus enhancing minor details while preserving generalization\cite{huang2017beyond}.
\subsubsection{Disentangle Representation}
Representation learning \cite{bengio2013representation} illustrated the way to synthesize new samples from a desired sample space. Deep learning encoder-decoder architecture has shown viable results in separating one attribute from others\cite{huang2007unsupervised}. There are cross-reconstruction based methods to disentangle identity and pose information \cite{Zhu_2013,Peng_2017}. We employ a similar concept in our two cyclic pathway architecture. We use weakly labeled data for representation learning, which makes the model more generalized and avoid overfitting. Native domain reconstruction using our cyclic pathway strategy preserves idiosyncratic information.

\section{Methodology}
Our objective is to learn a latent space which captures the pose independent identity information from the input face data. The learned space should be able to generate a new face sample, for a given identity and pose, having maximum perceptual information. In this section, we formulate the problem as a deep learning task following by explaining primary and secondary encoder-decoder network and then discussing the architecture training.

\subsection{Problem Formulation}
\begin{figure}
  \centering
  \includegraphics[width=\textwidth]{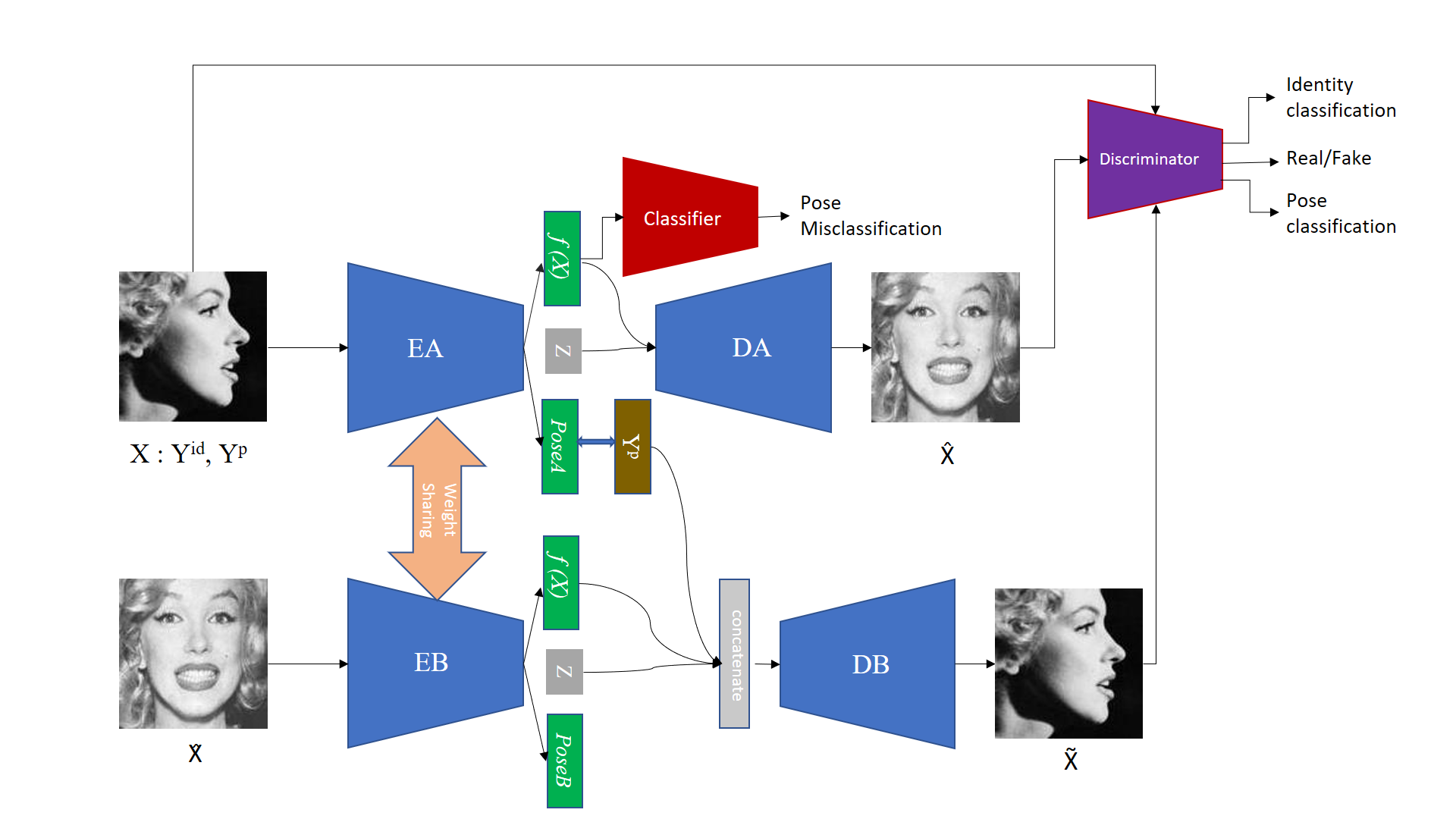}
  \caption{Proposed framework.}
  \label{fig1}
 \end{figure}

We symbolize input face image as X and corresponding pose and identity label as Y$^{id}$ and Y$^{p}$ respectively. The main objective is to synthesize a realistic face $\hat{X}$ for given pose and identity. We also use a secondary framework to generate another image $\tilde{X}$ from $\hat{X}$ which closely match the original image. We provide the primary framework with the input pairs (X: Y$^{id}$, Y$^{p}$) to create a manifold f(X) containing identity and pose apprehension as separate clusters. To synthesize a image $\hat{X}$,  f(X) can be traversed with given pose($\hat{Y}^{p}$) and identity($\hat{Y}^{id}$) input. We concatenate noise (Z) with f(X) and pose(P) to incorporate face details variations. The primary framework consists of encoder (EA), a latent classifier (LC) and decoder (DA). The secondary framework consists of encoder (EB) and decoder (DB). Both generative framework shares the common discriminator (D).\\
\noindent Mathematically this can be defined as:
\begin{equation} f(X) + poseA = EA(X: Y^{id}, Y^{p}) \end{equation}
\begin{equation} \hat{X} = DA(f(X), \hat{Y}^{p}, Z) \end{equation}
\begin{equation} f(X) + poseB = EB(\hat{X}) \end{equation}
\begin{equation} \tilde{X} = DB(f(X), Y^{p}, Z) \end{equation}
Figure \ref{fig1} shows visual representation of above equations.

\subsection{The Cyclic Framework}
We propose a cyclic GAN structure made from two back to back encoder-decoder framework named as the primary and secondary framework. The first framework aims to generate face for a given pose while the secondary framework seeks to create original domain image to minimize cyclic domain mismatch.
The primary framework consists of an encoder which projects the input data distribution to a new latent space and captures the necessary information which distinctly represents each input sample. While encoding the information, the latent encoding capture poses knowledge as well. We employ a binary classifier on latent encoding to make sure that latent encoding is separated from pose information. We train the classifier on actual pose labels so that if there is an extent of pose information present, then classifier will classify the latent encoding correctly. We train the encoder through the classifier gradient in such a way to produce latent encoding which doesn't have pose information and classifier miss classify it. This latent encoding is then passed to the decoder which is trained on discriminator gradient and produce photorealistic face images.\\
The secondary framework employs similar characteristics while sharing weights with the primary framework. The generated image from the primary framework, which is frontal in our case, is passed to the secondary framework. The second encoder, which works similarly to the first encoder, project the data into latent space. The latent encoding containing the identity information is concatenated with the true pose label to generate the original image.
The Discriminator gradient is propagated to the primary framework to adjust the latent representation.

In primary framework, we employ the encoder EA that compresses the input face image into a vector in latent space called latent vector. This latent vector inherits the pose as well as identity information from the input face image. We split the vector to distinguish pose and identity from each other.A binary classifier is employed to explicitly separate the pose. \textit{f(X)} is latent vector containing information about individual identities and \textit{pose} is latent vector containing information about face pose.

The labels are made based on frontal or profile view of the face. We create binary labels (0 and 1) for two kinds of input face image. First when the input face is frontal and second when the input face is not frontal. These labels are also used to train the binary classifier.

\subsection{Training the architecture}
\section{Experiments}
PIGAN intend to learn to disentangle representation as well as synthesize pose controlled face images. This section demonstrates the performance of both objectives. This sections also discuss the datasets, experimental setting, visual results, and comparison.
\subsection{Datasets and implementation settings}
We mostly performed various experiments on CFP (celebrity frontal profile) dataset. We also evaluate our framework on other widely used face datasets. CFP dataset is made up of 500 individuals each consists of 10 frontal face images and 4 profile face images. We train the initial models on 450 subjects and evaluate on remaining.
We follow \cite{Yi2014LearningFR} to pre-process the images.
We crop out all the images and align them in 100x100 because images section excluding face doesn't add much value for representation learning. As mentioned in \cite{Tran2017DisentangledRL}, we sample 96x96 part from the aligned images to create augmented data and normalize the image values in the range [-1,1]. The basic framework architecture follows the DC-GAN \cite{Radford2015UnsupervisedRL} implementation. For training purpose, batch size of 64 is used, and weights are initialized using the normal distribution of zero mean and 0.02 standard deviation. We use momentum optimizer (Adam) with 0.5 momentum and 0.0002 learning rate.

Learning rate is set to 0.0002 in order to provide stable training for GAN. Having high initial learning rate resulted in unstable training where one of the network (Generative or discriminative) learns faster and immobilize the learning for second one.

For training, we do not train generator and discriminator simultaneously. We hold the training for discriminator while the generator catches up and can fool the discriminator. This results in stable learning of the network, as mentioned in \cite{Goodfellow2014GenerativeAN}.
\subsection{Evaluation and Comparison}
We compare our proposed framework with DR-GAN \cite{tran2017disentangled}. We, following the DR-GAN, build upon the DC-GAN \cite{radford2015unsupervised} architecture to compare results on same stage. We train both the networks on same dataset and following the similar training strategy.

Among total of 500 subjects in CFP dataset we split the data into train to test ratio of 450:50 or 90:10.
\begin{table}
\centering
  \caption{Experimental results on CFP Dataset}
  \label{tab:freq}
  \begin{tabular}{ lll}
    \toprule
    \textbf{Method} & \textbf{Frontal-Frontal} &\textbf{Frontal-Profile}\\
	\midrule
    Sengupta et al.& 96.40 $\pm$ 0.69 & 84.91 $\pm$ 1.82\\
    DR-GAN & 97.13 $\pm$ 0.68 & 90.82 $\pm$ 0.28\\
    Current Work & 98.23 $\pm$ 0.83 & 91.92 $\pm$ 0.59\\
	\bottomrule
\end{tabular}
\end{table}

\subsubsection{Evaluation on IJB-A}
\begin{figure}
\centering
\includegraphics[width=\textwidth]{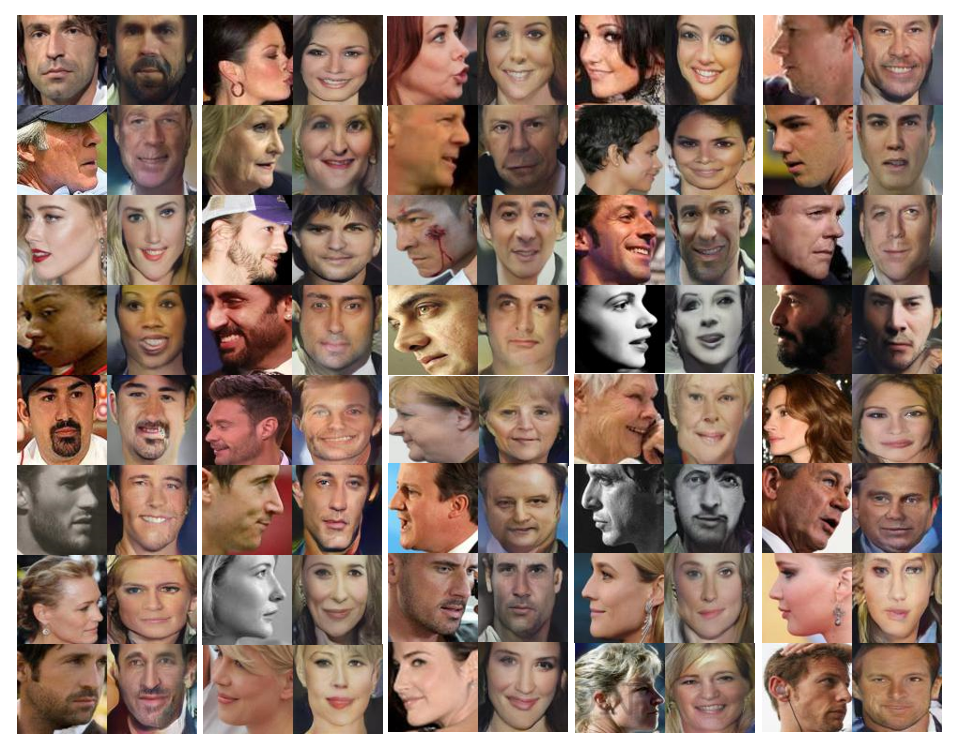}
\caption{Generated face images. Each pair shows left face as input and right face as generated frontal image.} \label{fig2}
\end{figure}

 We show the results of PI-GAN in figure \ref{fig2}, respectively. We compare it with DR-GAN, which produces images with sharpness, but the identities suffered. PI-GAN produces photo-realistic face images with high-quality. PI-GAN successfully synthesize good profile face images (Although not perfect enough), and DR-GAN fails to produce images from largely posed input faces.

\subsection{Visualization}
We use t-SNE \cite{Maaten2008VisualizingDU} for latent space visualization. We restrict the size of latent space to 256-dimensions and use t-SNE to project onto two dimensional. We observe the separability of different individuals and notice that large pose faces are clustered far away then their identity cluster.

GAN tries to capture the input data manifold to recreate the data (faces) it was trained on. As compared to classical machine learning, In deep learning features are described in terms of latent space. GAN uses a encoder to compress the input to a size of latent space which is then further used to generate similar faces.

\subsection{Results and Discussion}

We analyzed the generated faces and iterated over architecture to mitigate the shortcomings. In \ref{fig2}, we show the input image stitched together with generated image. While the primary framework disentangle the pose information, it also separates irrelevant information from the identity. This makes the overall generator robust to noisy data including occlusion, glasses, face injury etc. We observe the latent space manifold captures the idiosyncratic information about color and position of hair and beard which adds to better face generation having individual characteristics. This also makes the model invariant to illumination. The method disentangles the identity information from other information including pose as well as illumination which makes the method robust to pose and illumination. The model does not fail to synthesize frontal under large pose variance and high illumination difference.
Encoder does a great job in removing out noise from the test input like the identity of the person is not evident due to glasses or mask or some object occlusion, In that case we need to supply more images of the same person to collect as much information. The intra-class robustness can also be handled by having more test images for the person.
We preprocess the input images by centering and cropping out the unnecessary borders. To tackle the issue of affine in-variance, model should be trained on affine transformed images which can be done while preprocessing the data and it aids in creating more training data.

\subsection{Future direction and Applications}

The existing literature in this area consist of deep learning networks having single GAN. We contribute by adding dual GAN architecture for synthesizing frontal faces. As evident one of the application of this research falls in security where one can use CCTV camera frames to extract profile images and generate frontal face from them to identify the targeted person. We provide a concept of disentangling two  information based on provided labels. This is not only useful in case of human face but for animals, objects and other things.
Multiple application can be built on this, from rotating a face, removing occlusion to generating unseen faces.A future work in this area may include adding more robustness to the architecture by improving affine transformation. Also, exploring the areas like object synthesizing where this approach can be used.

\section{Conclusion}
In this paper, We present a cyclic framework for learning pose-independent identity representation. We use this representation for synthesizing photo-realistic images of faces for given pose and identity. The back-to-back encoder-decoder architecture consists of primary and secondary networks. The primary network creates the latent distribution and secondary network provide loss gradients to adjust the distribution to generate more realistic and identity preserving image. We observe that using a secondary framework allow us to employ direct losses between original and generated image at the pixel level, which was difficult using the primary framework alone due to viewpoint variation. We use weight sharing between two frameworks in order to project the input image on the same latent space. A latent classifier is used in the primary framework, which explicitly disentangles the pose information from identity information by intentionally performing pose miss-classification. Experiment results demonstrate the performance of the proposed framework, visually showing frontal faces. Our framework leverages the disentanglement representation as well as cyclic loss to enhance the perceptual quality.

The presented work is not is not only limited to frontalizing profile faces but also applicable for generating unseen faces, varying face attributes and applying the disentanglement concept for other objects.

\bibliographystyle{unsrt}
\bibliography{references}  

\end{document}